\documentclass[twoside,11pt]{article}

\usepackage{blindtext}

\usepackage[abbrvbib, preprint]{jmlr2e}

\usepackage{lastpage}
\usepackage{algorithm}
\usepackage{algpseudocode}
\usepackage{svg}
\usepackage{listings}
\usepackage{color}
\hypersetup{colorlinks, linkcolor=red, citecolor = blue}

\definecolor{dkgreen}{rgb}{0,0.6,0}
\definecolor{gray}{rgb}{0.5,0.5,0.5}
\definecolor{mauve}{rgb}{0.58,0,0.82}

\ShortHeadings{}{Musgrave et al.}
\firstpageno{1}

\begin{document}

\title{\raisebox{-0.3ex}{\includegraphics[scale=0.09]{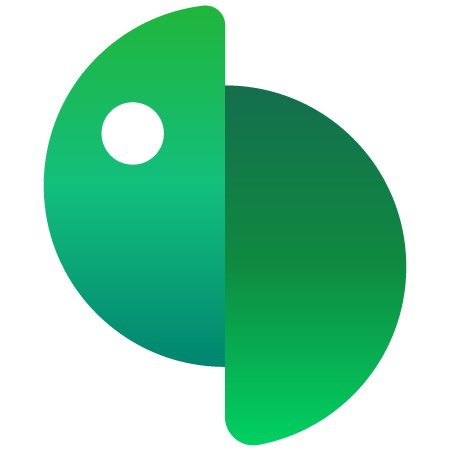}} \kern 0.18em PyTorch Adapt}

\author{\name Kevin Musgrave \\
\addr Cornell Tech \\
\AND
\name Serge Belongie \\
\addr University of Copenhagen \\
\AND
\name Ser-Nam Lim \\
\addr Meta AI
}

\editor{}

\maketitle

\begin{abstract}
PyTorch Adapt is a library for domain adaptation, a type of machine learning algorithm that re-purposes existing models to work in new domains. It is a fully-featured toolkit, allowing users to create a complete train/test pipeline in a few lines of code. It is also modular, so users can import just the parts they need, and not worry about being locked into a framework. One defining feature of this library is its customizability. In particular, complex training algorithms can be easily modified and combined, thanks to a system of composable, lazily-evaluated hooks. In this technical report, we explain in detail these features and the overall design of the library. Code is available at \href{https://www.github.com/KevinMusgrave/pytorch-adapt}{github.com/KevinMusgrave/pytorch-adapt}.
\end{abstract}

\section{Design}

\begin{figure}[H]
\centering
\includegraphics[width=0.75\textwidth]{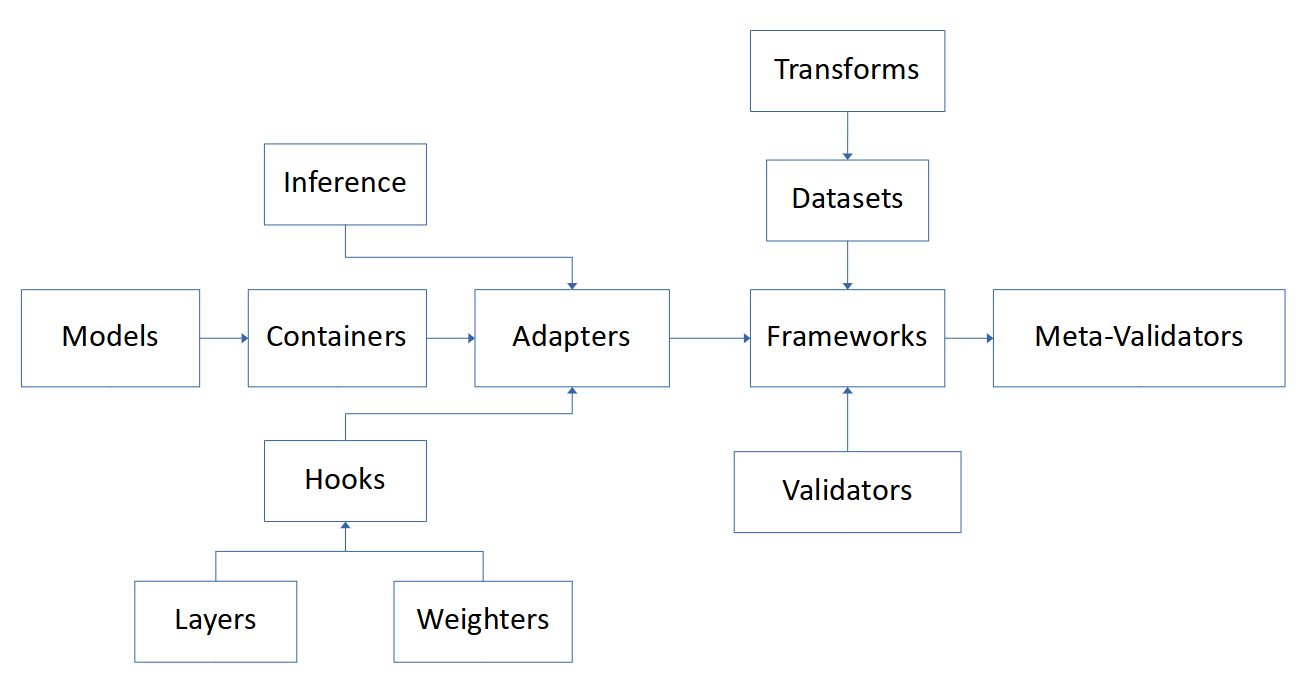}
\caption{How the modules are related to each other. An arrow going from X $\rightarrow$ Y indicates that Y entities receive X entities as input.}
\label{design:ModulesOverview}
\end{figure}

Figure \ref{design:ModulesOverview} shows how the library's 12 modules are used by one another. In the following sections, we explain the purpose of the most important modules.

\subsection{Hooks}
Domain adaptation algorithms come in many forms. Some use a GAN architecture. Some require features from one domain, but not the other. Some require features to be detached from the autograd graph, and so on. This makes it difficult to design code that:

\begin{itemize}
    \item is easy to customize
    \item does not perform unnecessary computations
\end{itemize}

\begin{figure}[H]
\centering
\includegraphics[width=0.71\textwidth]{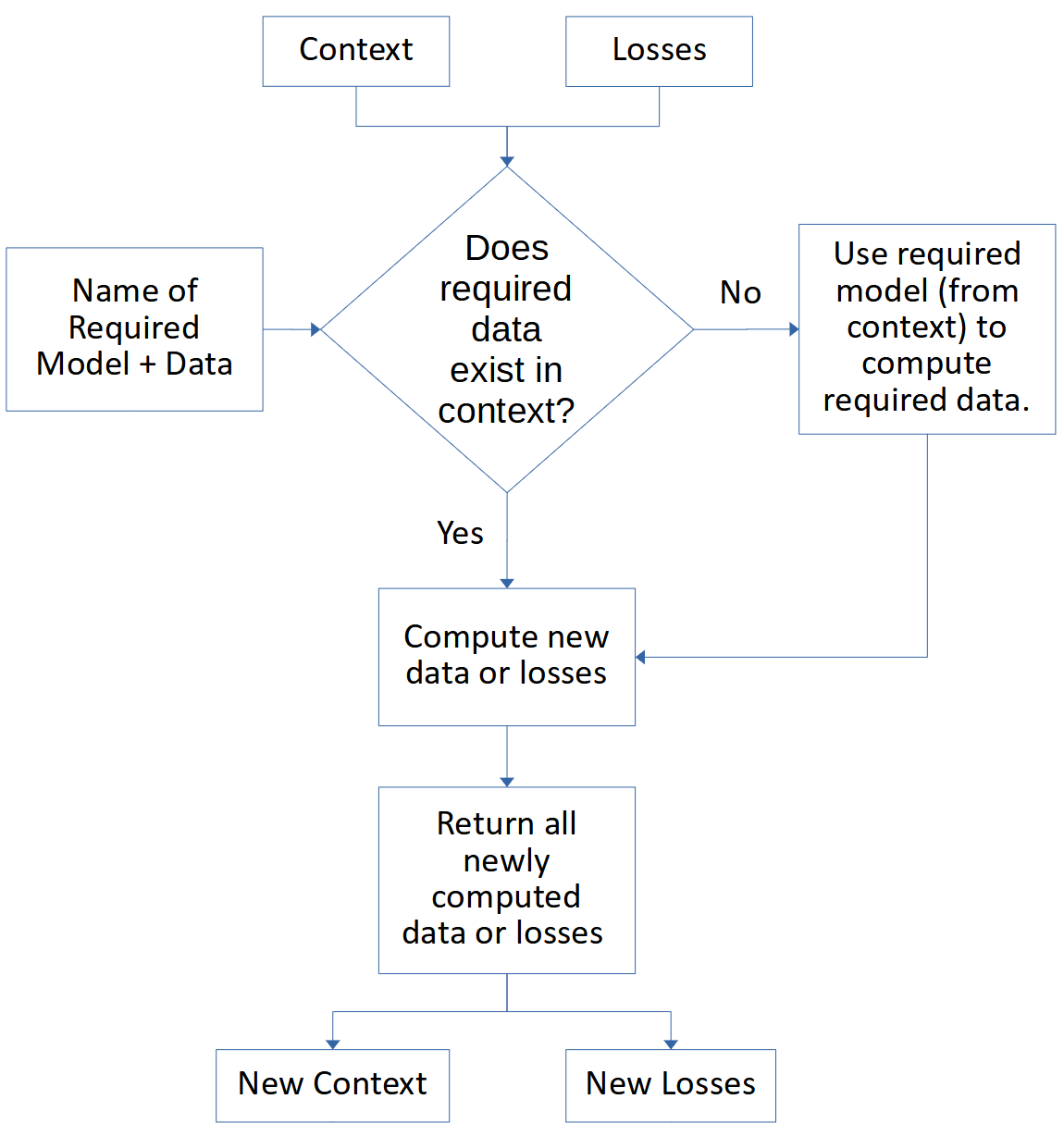}
\caption{The overall control flow of hooks.}
\label{design:HookAlgorithm}
\end{figure}

\noindent To solve this problem, PyTorch Adapt uses composable, lazily evaluated hooks. In a nutshell, each hook follows these steps (see also Figure \ref{design:HookAlgorithm}):

\begin{enumerate}
\item Determine the required data (e.g. source features, target logits).
\item Check if the required data is already available in the current context. If not, then compute it.
\item Use the required data.
\end{enumerate}

\noindent The ``context" is a Python dictionary mapping from strings to arbitrary values (e.g. PyTorch models and tensors). A hook's required data is specified using one of the various ``features" hooks. In the example below, the \texttt{FeaturesHook} checks if \texttt{"src\_imgs\_features"} is available in the context. If not, then it uses the \texttt{"G"} model to compute it.

\begin{lstlisting}
from pytorch_adapt.hooks import BaseHook, FeaturesHook
from pytorch_adapt.utils import common_functions as c_f

class SimpleHook(BaseHook):
    def __init__(self, **kwargs):
        super().__init__(**kwargs)
        self.loss_name = "L2NormLoss"
        self.f_hook = FeaturesHook(model_name="G", domains=["src"])

    def call(self, inputs, losses):
        outputs = self.f_hook(inputs, losses)[0]
        features = c_f.extract([outputs, inputs], ["src_imgs_features"])[0]
        loss = torch.norm(features)
        return outputs, {self.loss_name: loss}

    def _loss_keys(self):
        return [self.loss_name]

    def _out_keys(self):
        return self.f_hook.out_keys
\end{lstlisting}

\noindent Hooks are composable. For example, the \texttt{ChainHook} allows users to chain together an arbitrary number of hooks and share context between them. The hooks are run sequentially, with the outputs of hook $n$ being added to the context so that they become part of the inputs to hook $n+1$.

\begin{lstlisting}
from pytorch_adapt.hooks import ChainHook

hook = ChainHook(hook1, hook2, hook3)
# In the following call, hook3 will receive initial_context
# plus all new data computed by hook1 and hook2
outputs, losses = hook(initial_context)
\end{lstlisting}

\noindent Hooks allow algorithms to be easily modified, without causing unnecessary computations. In the example below, \texttt{hook1} is a \texttt{ClassifierHook}, which applies a cross-entropy loss to the source domain logits. This hook does not require target domain features or logits, hence they are not computed. On the other hand, \texttt{hook2} is the same \texttt{ClassifierHook} but with the addition of \texttt{BSPHook} and \texttt{BNMHook}. These additional hooks do require target features and logits, so they \textit{will} be computed for \texttt{hook2}. 
\begin{lstlisting}
from pytorch_adapt.hooks import BNMHook, BSPHook, ClassifierHook

hook1 = ClassifierHook(optimizers)
hook2 = ClassifierHook(optimizers, post=[BSPHook(), BNMHook()])
\end{lstlisting}

\subsection{Adapters, Containers, and Inference Functions}
The \texttt{adapters} module contains thin wrappers around hooks and inference functions.  

\begin{lstlisting}
from pytorch_adapt.adapters import DANN

adapter = DANN(models)
# During training
loss = adapter.training_step(data)
# During evaluation
output = adapter.inference(data)
\end{lstlisting}

\noindent One of the benefits of adapters is simplified object manipulation via \texttt{containers}. Containers are Python dictionaries with additional functions that allow for programmatic object creation:
\begin{lstlisting}
from pytorch_adapt.containers import Models, Optimizers, LRSchedulers

G = torch.nn.Linear(1000, 100)
C = torch.nn.Linear(100, 10)

models = Models({"G": G, "C": C})
optimizers = Optimizers((Adam, {"lr": 0.456, "weight_decay": 0.123}))
schedulers = LRSchedulers((ExponentialLR, {"gamma": 0.99}))

# internally creates an optimizer for each model
# and a scheduler for each optimizer
adapter = DANN(models, optimizers, schedulers) 
\end{lstlisting}

\noindent Domain adaptation algorithms vary in structure, and therefore can also vary in how inference is performed. Some structures allow for multiple ways of performing inference. Consider MCD, which contains multiple classifier branches. A user may want the sum of the branches logits, or the logits from each branch separately. The \texttt{inference} module provides functions that allow adapters to be customized in this way.

\begin{lstlisting}
from pytorch_adapt.adapters import MCD
from pytorch_adapt.inference import mcd_fn, mcd_full_fn

# returns features and logits during inference
adapter1 = MCD(models=models, inference_fn=mcd_fn)
# returns features, logits, logits1, and logits2 during inference
adapter2 = MCD(models=models, inference_fn=mcd_full_fn)
\end{lstlisting}

\subsection{Frameworks and Validators}
The \texttt{frameworks} module provides classes for integrating adapters into existing frameworks, like PyTorch Ignite \citep{pytorch-ignite}.

\newpage
\begin{lstlisting}
from pytorch_adapt.frameworks.ignite import Ignite

trainer = Ignite(adapter)
trainer.run(dataloaders)
\end{lstlisting}

\noindent Usually an evaluation metric is needed during training, to determine which checkpoint is best. In the supervised setting, this is a trivial task as accuracy can be computed directly based on labels. However, in the unsupervised setting, labels are not available, and evaluation becomes much more difficult. In fact, this is an active area of research. The \texttt{validators} module provides numerous validation methods to compute or estimate accuracy with or without access to labels.

\begin{lstlisting}
from pytorch_adapt.validators import BNMValidator

validator = BNMValidator()
# use independently
score = validator(target_train={"logits": logits})

# or with a framework wrapper
trainer = Ignite(adapter, validator)
best_score, best_epoch = trainer.run(dataloaders)
\end{lstlisting}

\section{Related Libraries}

Other open-source domain adaptation libraries include Dassl \citep{dassl}, Transfer-Learning-Library \citep{tllib}, and ADAPT \citep{de2021adapt}. PyTorch Adapt's main differentiating feature is its emphasis on customizability.

\section{Acknowledgements}

Thanks to Jeff Musgrave for designing the logo.

\vskip 0.2in
\bibliography{references}

\end{document}